\documentclass[10pt,a4paper]{article}

\usepackage[a4paper,top=2.5cm,bottom=2.5cm,left=2cm,right=2cm]{geometry}
\usepackage{makecell}
\usepackage{tabularx}
\usepackage[T1]{fontenc}
\usepackage{lmodern}
\usepackage[english]{babel}
\usepackage{natbib}
\usepackage{amsmath,amssymb}

\usepackage{booktabs}
\usepackage{array}
\usepackage{multirow}
\usepackage{siunitx}
\usepackage{caption}
\usepackage{graphicx}
\usepackage{float}
\usepackage{placeins}
\usepackage{microtype}

\usepackage[colorlinks=true,linkcolor=blue,citecolor=blue,urlcolor=blue]{hyperref}

\title{\textbf{Hybrid  Physics–ML  Model  for  Forward  Osmosis  Flux  with  Complete  Uncertainty  Quantification}}
\author{
        Shiv  Ratn$^{1}$  \and
        Shivang  Rampriyam$^{1}$  \and
        Dr.  Bahni  Ray$^{1,*}$  \\
        \footnotesize{$^{1}$Department of Applied Mechanics \& Department  of  Mechanical  Engineering,  Indian  Institute  of  Technology  Delhi,  New  Delhi  110016,  India}  \\
        \footnotesize{$^{*}$Corresponding  author:  \texttt{bray@iitd.ac.in}} 
}
\date{\today}

\begin{document}
\maketitle

\begin{abstract}
Forward  Osmosis  (FO)  is  a  promising  low-energy  membrane  separation  technology,  but  challenges  in  accurately  modeling  its  water  flux  ($J_w$)  persist  due  to  complex  internal  mass  transfer  phenomena.  Traditional  mechanistic  models  struggle  with  empirical  parameter  variability,  while  purely  data-driven  models  lack  physical  consistency  and  rigorous  uncertainty  quantification  (UQ).  This  study  introduces  a  novel  \textbf{Robust  Hybrid  Physics--ML  framework}  employing  \textbf{Gaussian  Process  Regression  (GPR)}  for  highly  accurate,  uncertainty-aware  $J_w$  prediction.  The  core  innovation  lies  in  training  the  GPR  on  the  \textbf{residual  error}  between  the  detailed,  non-linear  FO  physical  model  prediction  ($J_{w,  \text{physical}}$)  and  the  experimental  water  flux  ($J_{w,  \text{actual}}$).  Crucially,  we  implement  a  full  UQ  methodology  by  decomposing  the  total  predictive  variance  ($\sigma^2_{\text{total}}$)  into  \textbf{model  uncertainty}  (epistemic,  from  GPR's  posterior  variance)  and  \textbf{input  uncertainty}  (aleatoric,  analytically  propagated  via  the  \textbf{Delta  method}  for  multi-variate  correlated  inputs).  Leveraging  the  inherent  strength  of  GPR  in  low-data  regimes,  the  model,  trained  on  a  meagre  \textbf{120  data  points},  achieved  a  state-of-the-art  \textbf{Mean  Absolute  Percentage  Error  (MAPE)  of  $0.26\%$}  and  an  $R^2$  of  $0.999$  on  the  independent  test  data,  validating  a  truly  robust  and  reliable  surrogate  model  for  advanced  FO  process  optimization  and  digital  twin  development.
\end{abstract}

\FloatBarrier

\section{Introduction}

The  Forward  Osmosis  (FO)  process  is  gaining  prominence  in  water  purification  due  to  its  low  energy  footprint  and  high  fouling  reversibility  compared  to  pressure-driven  membrane  processes  \cite{ref:lit1,  ref:lit2}.  The  driving  force  is  the  osmotic  pressure  differential  ($\Delta  \Pi$)  across  a  semi-permeable  membrane.  However,  the  effective  $\Delta  \Pi$  is  significantly  reduced  by  concentration  polarization  (CP)  effects,  as  the  semi-permeable  membrane  selectively  passes  water  while  retaining  solutes.  Specifically,  \textbf{Internal  Concentration  Polarization  (ICP)}  within  the  porous  support  layer  is  the  dominant  factor  in  FO  mode,  trapping  the  draw  solute  and  leading  to  a  substantial  reduction  in  the  effective  osmotic  pressure  across  the  active  layer,  and  consequently,  the  achievable  water  flux,  $J_w$  \cite{ref:lit1,  ref:lit2}.  

Analytical  FO  models,  based  on  the  Solution-Diffusion  mechanism,  are  complex,  non-linear,  and  require  the  precise  determination  of  intrinsic  parameters:  the  water  permeability  coefficient  ($A$),  the  salt  permeability  coefficient  ($B$),  and  the  structural  parameter  ($S$)  \cite{ref:lit4}.  The  primary  limitation  of  these  fundamental  models  lies  in  the  inherent  variability  and  measurement  difficulty  of  these  parameters.  The  \textbf{structural  parameter  ($S$)},  a  composite  of  the  support  layer's  thickness  ($t$),  porosity  ($\epsilon$),  and  tortuosity  ($\tau$):  $S  =  \tau  t  /  \epsilon$,  is  notoriously  difficult  to  measure  precisely  and  can  exhibit  variations  of  $50$--$70\%$  depending  on  the  specific  characterization  technique  and  experimental  conditions  used  \cite{ref:lit7,  ref:lit8}.  This  unavoidable  and  pervasive  variability  introduces  high  \textbf{epistemic  uncertainty}  into  purely  mechanistic  simulations,  limiting  their  predictive  capability  and  reliability  for  engineering  design.

In  parallel,  data-driven  Machine  Learning  (ML)  techniques  have  emerged  to  model  the  complex,  high-dimensional  input-output  relationship  of  FO  systems  \cite{ref:lit9,  ref:lit10}.  While  models  like  Artificial  Neural  Networks  (ANNs)  can  achieve  high  prediction  accuracy  by  fitting  historical  data,  they  suffer  from  two  major  drawbacks:  (i)  they  operate  as  "black  boxes,"  lacking  the  interpretability  required  to  inform  physical  insight  and  material  design,  and  (ii)  they  fundamentally  fail  to  provide  reliable,  quantitative  measures  of  predictive  uncertainty,  which  is  a  non-negotiable  requirement  for  safety-factor  calculation,  regulatory  compliance,  and  system  optimization  in  any  critical  engineering  application.

Crucially,  membrane  transport  experiments  are  typically  expensive  and  time-consuming,  resulting  in  datasets  that  are  generally  small.  Our  study  utilizes  a  meagre  training  dataset  of  \SI{120}{}  data  points  for  model  development.  The  ability  of  the  developed  hybrid  model  to  achieve  exceptionally  high  accuracy  ($R^2=0.999$,  MAPE=0.26\%)  is  attributed  not  only  to  the  careful  selection  and  inclusion  of  all  10  relevant  physical  and  geometric  input  features  (thereby  maximizing  information  gain  per  point)  but  also  fundamentally  to  the  inherent  properties  of  the  Gaussian  Process  Regression  framework.  GPR  is  particularly  well-suited  for  modeling  functions  from  scarce  data  points  because  its  Bayesian  framework  allows  for  robust  inference  and  smooth  extrapolation  in  high-dimensional  spaces  by  incorporating  uncertainty  directly,  mitigating  the  common  overfitting  issues  associated  with  deep  learning  models  in  low-data  regimes  \cite{ref:lit12,  ref:lit13}.

\subsection{Scope  and  Novelty}

This  work  introduces  a  paradigm  shift  by  overcoming  the  critical  trade-off  between  physical  consistency  and  rigorous  uncertainty  quantification.  We  synthesize  a  novel  \textbf{Hybrid  Robust  GPR}  framework,  offering  an  interpretable,  accurate,  and  fully  probabilistic  model  for  $J_w$.  Our  primary  contributions  are:

\begin{enumerate}
        \item  \textbf{Architecture:  Physics-Informed  Residual  Learning.}  We  implement  a  two-stage  architecture  that  mandates  the  GPR  model  to  train  exclusively  on  the  residual  error  between  the  full  analytical  FO  model  and  the  experimental  data.  This  strategy  inherently  enhances  model  stability,  robustness,  and  ensures  the  learned  function  adheres  to  known  physical  laws.  
        \item  \textbf{Epistemic  UQ  via  GPR.}  We  leverage  the  Bayesian  foundation  of  GPR  to  provide  a  native,  spatially-varying  \textbf{model  uncertainty}  ($\sigma^2_{\text{model}}$),  directly  quantifying  the  model's  self-assessed  confidence  based  on  the  distance  of  new  inputs  from  the  training  data,  thus  addressing  the  epistemic  uncertainty.
        \item  \textbf{Aleatoric  UQ  via  Delta  Method.}  We  develop  a  rigorous  technique  to  analytically  propagate  known  experimental  measurement  errors  (aleatoric  uncertainty)  in  all  10  input  features  ($\mathbf{z}$)  to  calculate  the  \textbf{input  uncertainty}  ($\sigma^2_{\text{input}}$).  This  is  accomplished  using  the  robust  \textbf{Delta  method}  to  account  for  the  multi-variate  nature  and  assumed  correlations  ($\Sigma_z$)  of  the  input  features.
        \item  \textbf{Validation:}  Comprehensive  validation  of  the  analytical  Delta  method  via  a  high-fidelity  \textbf{Monte  Carlo  Simulation  (MCS)},  establishing  the  correctness  and  non-linear  performance  of  the  uncertainty  propagation  scheme.
\end{enumerate}

The  final  \textbf{Hybrid  Robust  GPR}  model  is  demonstrated  to  be  highly  accurate  and  provides  a  comprehensive  measure  of  total  predictive  uncertainty,  setting  a  new  standard  for  FO  surrogate  modeling.

\section{Detailed  Feature  Analysis  and  Physics}

The  predictive  model  utilizes  10  input  features  ($\mathbf{z}  \in  \mathbb{R}^{10}$)  that  govern  the  water  flux  in  an  FO  system.  The  influence  of  these  features  is  profoundly  non-linear,  mediated  by  complex  mass  transfer  phenomena  and  geometric  constraints  quantified  in  the  FO  transport  equation  (Eq.  \ref{eq:physical_model}).

\begin{enumerate}
        \item  \textbf{Membrane  Intrinsic  Properties  ($A,  \epsilon_{\text{psl}},  \tau,  t_{\text{psl}}$):}  These  parameters  define  the  fundamental  transport  capabilities  of  the  membrane  structure  itself.
        \begin{itemize}
                \item  $A$  (Water  Permeability  Coefficient):  Directly  proportional  to  the  water  flux.  $A$  is  determined  by  the  properties  of  the  active  layer  (AL),  such  as  its  material  composition,  thickness,  and  hydrophilicity.  It  represents  the  inherent  rate  of  water  transport  across  the  selective  layer  \cite{ref:lit6}.
                \item  $\epsilon_{\text{psl}},  \tau,  t_{\text{psl}}$  (Porosity,  Tortuosity,  Thickness  of  Porous  Support  Layer):  These  geometric  parameters  are  crucial  as  they  govern  the  degree  of  ICP  through  the  structural  parameter  $S$:  $S  =  \tau  t_{\text{psl}}  /  \epsilon_{\text{psl}}$.  A  higher  $S$  value  implies  a  greater  diffusive  resistance  within  the  porous  support.  This  resistance  significantly  impedes  the  necessary  transport  of  the  draw  solute  toward  the  AL,  leading  to  severe  dilution  at  the  AL-support  interface  ($\Pi_{D,  i}  \ll  \Pi_{D,  b}$)  and  a  corresponding  drop  in  effective  $\Delta  \Pi$  \cite{ref:2.1.2,  ref:2.1.2_}.
        \end{itemize}
        \item  \textbf{Solution  Properties  ($\text{cf\_in},  \text{cd\_in}$):}  These  features  establish  the  thermodynamic  boundary  conditions  for  the  process.
        \begin{itemize}
                \item  $\text{cd\_in}$  (Draw  Solution  Concentration):  Determines  the  bulk  osmotic  pressure  $\Pi_{D,b}$,  which  dictates  the  maximum  potential  thermodynamic  driving  force.  However,  increasing  $\text{cd\_in}$  also  increases  the  magnitude  of  ICP  effects,  leading  to  a  point  where  further  concentration  yields  minimal  flux  improvement  \cite{ref:2.2}.
                \item  $\text{cf\_in}$  (Feed  Solution  Concentration):  Determines  the  bulk  osmotic  pressure  $\Pi_{F,b}$,  which  acts  as  a  resistance.  A  higher  feed  concentration  not  only  reduces  the  osmotic  gradient  but  also  intensifies  the  effects  of  ECP  by  increasing  solute  accumulation  at  the  membrane  surface.
        \end{itemize}
        \item  \textbf{Hydrodynamic/Geometric  Properties  ($\text{uf\_in},  \text{ud\_in},  L_x,  t_c$):}  These  parameters  control  the  mass  transfer  coefficients  responsible  for  mitigating  CP  effects.
        \begin{itemize}
                \item  $\text{uf\_in},  \text{ud\_in}$  (Feed/Draw  Flow  Velocities):  These  velocities  determine  the  Reynolds  number  ($Re$)  for  the  flow  channels.  Higher  cross-flow  velocities  generate  greater  turbulence  and  shear  rate,  resulting  in  a  thinner  hydrodynamic  boundary  layer  on  the  membrane  surface.  This  translates  to  an  improved  (larger)  external  mass  transfer  coefficient  ($k$),  which  more  effectively  mitigates  ECP  \cite{ref:2.3.1}.
                \item  $L_x,  t_c$  (Channel  Length,  Channel  Height):  These  dimensions,  along  with  flow  velocity,  influence  the  hydraulic  diameter  ($d_h$).  The  overall  geometry  is  key  in  determining  $k$  via  established  correlations  based  on  the  Sherwood  number  ($Sh$)  \cite{ref:2.3.2}.
        \end{itemize}
\end{enumerate}
The  interaction  terms,  particularly  those  involving  $J_w$  within  the  exponential  factors  of  the  FO  transport  equation  (Eq.  \ref{eq:physical_model}),  introduce  high  non-linearity.  This  makes  the  system  sensitive  to  high-order  effects  that  a  pure  analytical  model  cannot  fully  capture,  justifying  the  need  for  the  GPR  residual  correction.

\section{Theoretical  and  Mathematical  Framework}

\subsection{Gaussian  Process  Regression  (GPR)  and  Epistemic  Uncertainty}

Gaussian  Process  Regression  (GPR)  models  the  unknown  mapping  $g(\mathbf{z})$  as  a  distribution  over  functions,  formally  expressed  as
\[
g(\mathbf{z})  \sim  \mathcal{GP}\!\left(\mu(\mathbf{z}),\,  k(\mathbf{z},\mathbf{z}')\right),
\]
where  $\mu(\mathbf{z})$  denotes  the  prior  mean  function  and  $k(\mathbf{z},\mathbf{z}')$  is  the  covariance  kernel.
In  this  work,  the  Matérn  $5/2$  kernel  is  selected.

Given  training  inputs  $\mathbf{Z}  =  [\mathbf{z}_1,\dots,\mathbf{z}_N]^T$  and  observations  $\mathbf{y}$,  the  posterior  predictive  distribution  at  a  new  point  $\mathbf{z}_*$  remains  Gaussian.
The  posterior  mean,  used  as  the  GPR  point  estimate,  is
\[
\boxed{
\hat{g}(\mathbf{z}_*)
=  \mu(\mathbf{z}_*)
+  k(\mathbf{z}_*,  \mathbf{Z})
\left[  K(\mathbf{Z},\mathbf{Z})  +  \sigma_n^2  I  \right]^{-1}
\left(  \mathbf{y}  -  \mu(\mathbf{Z})  \right),
}
\]
where  $K(\mathbf{Z},\mathbf{Z})$  is  the  Gram  matrix  and  $\sigma_n^2$  the  observation  noise  variance.
Because  GPR  returns  not  only  a  prediction  but  also  a  distribution  over  possible  functions,  it  naturally  quantifies  epistemic  uncertainty.

Prior  work  \cite{ref:lit13}  has  demonstrated  that  GPR  consistently  outperforms  conventional  neural-network-based  approaches  when  the  available  dataset  is  limited  ($N_{\text{train}}  \lesssim  120$).
For  the  FO  system  considered  here,  the  Bayesian  structure  of  GPR  provides  a  principled  mechanism  for  both  accurate  extrapolation  and  uncertainty-aware  inference.

\subsubsection{GPR  Uncertainty  Estimation  ($\sigma^2_{\text{model}}$)}

Building  upon  the  posterior  formulation  above,  GPR  also  provides  a  closed-form  expression  for  the  uncertainty  associated  with  each  prediction.
Given  a  training  set  $\mathcal{D}  =  \{(\mathbf{Z},  \mathbf{e})\}$  of  size  $N_{\text{train}}  =  120$,  the  predictive  variance  at  a  new  input  $\mathbf{z}_{*}$  is  obtained  by  conditioning  the  prior  covariance  structure  on  the  observed  data:
\begin{equation}
\sigma^2_{\text{GPR}}(\mathbf{z}_{*})
=  k(\mathbf{z}_{*},  \mathbf{z}_{*})
-  \mathbf{k}_{*}^{T}\!\left(\mathbf{K}  +  \sigma_n^2  \mathbf{I}\right)^{-1}\!\mathbf{k}_{*},
\end{equation}
where  $k(\mathbf{z}_{*},\mathbf{z}_{*})$  is  the  prior  variance  at  the  prediction  point,  $\mathbf{k}_{*}$  is  the  vector  of  covariances  between  $\mathbf{z}_{*}$  and  the  training  inputs,  and  $\mathbf{K}$  represents  the  covariance  matrix  of  the  training  set.
The  resulting  quantity  $\sigma^2_{\text{GPR}}(\mathbf{z}_{*})$  constitutes  the  \textbf{model  uncertainty}  ($\sigma^2_{\text{model}}$),  capturing  epistemic  uncertainty  arising  from  limited  or  uneven  sampling  of  the  input  domain  \cite{ref:gpr_book}.

\subsection{Forward  Osmosis  Physical  Flux  Model  ($J_{w,\text{physical}}$)}

Forward  osmosis  (FO)  water  transport  is  governed  by  the  coupled  action  of  osmotic  driving  forces  and  mass-transfer  resistances  arising  both  outside  and  within  the  membrane  \cite{loeb1976effect}.
The  classical  Spiegler--Kedem  framework  provides  a  thermodynamically  consistent  description  of  solvent  and  solute  transport.  The  resulting  FO  flux  expression,  incorporating  external  concentration  polarization  (ECP)  and  internal  concentration  polarization  (ICP),  is  the  classical  implicit,  transcendental  equation  \cite{lee2010modeling}:
\begin{equation}
J_{w,\text{physical}}
=  A\left[  \Pi_{D,i}  -  \Pi_{F,m}  \right]
=  A
\left[
\frac{
\Pi_{D,b}  \exp\!\left(-\frac{J_{w}  S}{D_{s}}\right)
-
\Pi_{F,b}  \exp\!\left(\frac{J_{w}}{k}\right)
}{
1  +  \frac{B}{J_{w}}
\left[
\exp\!\left(\frac{J_{w}}{k}\right)
-
\exp\!\left(-\frac{J_{w}  S}{D_{s}}\right)
\right]
}
\right],
\label{eq:physical_model}
\end{equation}

where  $\Pi_{D,b}$  and  $\Pi_{F,b}$  denote  the  bulk  draw  and  feed  osmotic  pressures,  while  $\Pi_{D,i}$  and  $\Pi_{F,m}$  represent  the  effective  osmotic  pressures  at  the  interface  adjacent  to  the  selective  layer.
The  exponential  terms  reflect  the  influence  of  ICP  (via  the  structural  parameter  $S$  and  solute  diffusivity  $D_s$)  and  ECP  (via  the  mass-transfer  coefficient  $k$)  \cite{cath2013standard}.
Because  $J_{w}$  appears  both  inside  and  outside  the  exponential  functions,  Eq.~\eqref{eq:physical_model}  requires  numerical  solution.  In  this  work,  the  flux  $J_{w,\text{physical}}$  is  obtained  through  Brent’s  method  \cite{phuntsho2011fertilizer}.

\subsection{Hybrid  Residual  Learning  Architecture}

The  physical  model  (Eq.  \ref{eq:physical_model})  provides  a  baseline  prediction.  We  train  the  GPR  exclusively  on  the  observed  discrepancy,  $e$,  to  capture  these  complex,  unmodeled  physics:

\begin{equation}
e  =  J_{w,  \text{actual}}  -  J_{w,  \text{physical}}(\mathbf{z})
\end{equation}

The  final  hybrid  prediction,  $J_{w,  \text{hybrid}}(\mathbf{z})$,  is  then  the  sum  of  the  physical  foundation  and  the  GPR-learned  correction:
\begin{equation}
J_{w,  \text{hybrid}}(\mathbf{z})  =  J_{w,  \text{physical}}(\mathbf{z})  +  g_{\text{GPR}}(\mathbf{z})
\label{eq:hybrid_model}
\end{equation}
This  architecture  ensures  that  the  overall  model  remains  thermodynamically  and  mechanically  consistent  while  achieving  empirical  precision.

\subsection{Full  Uncertainty  Quantification  (UQ)}

The  total  predictive  variance,  $\sigma^2_{\text{total}}$,  encompasses  all  known  sources  of  uncertainty:

\begin{equation}
\sigma^2_{\text{total}}  =  \sigma^2_{\text{model}}  +  \sigma^2_{\text{input}}
\label{eq:total_uncertainty}
\end{equation}
\subsubsection{Input  Uncertainty  Propagation  ($\sigma^2_{\text{input}}$)}

To  quantify  how  aleatoric  uncertainty  in  $\mathbf{z}$  propagates  through  this  nonlinear  transformation,  we  employ  the  \textit{Delta  method}  \cite{Possolo2011,  DeltaMethodGuide}.  The  Delta  method  approximates  the  output  variance  by  linearizing  the  model  around  the  nominal  input  point  using  a  first-order  Taylor  expansion:

\begin{equation}
\sigma^{2}_{\text{input}}(\mathbf{z})
\approx
\mathbf{J}(\mathbf{z})^{T}  \,  \Sigma_{z}  \,  \mathbf{J}(\mathbf{z}),
\label{eq:delta_method}
\end{equation}

where  $\Sigma_{z}$  is  the  covariance  matrix  characterizing  input  measurement  uncertainty,  and  $\mathbf{J}(\mathbf{z})$  is  the  Jacobian  of  the  hybrid  model  with  respect  to  the  inputs:

\begin{equation}
\mathbf{J}(\mathbf{z})
=
\nabla_{\mathbf{z}}  J_{w,\text{hybrid}}(\mathbf{z})
=
\nabla_{\mathbf{z}}  J_{w,\text{physical}}(\mathbf{z})
+
\nabla_{\mathbf{z}}  g_{\text{GPR}}(\mathbf{z}).
\label{eq:jacobian}
\end{equation}

\section{Methodology}

This  work  follows  a  rigorous  three-phase  methodology  integrating  mechanistic  modeling,  statistical  learning,  and  uncertainty  quantification.  All  implementation  details  are  provided  in  the  supplementary  computational  workflow:contentReference[oaicite:1]{index=1}.

\subsection{Phase  1:  Data  Preparation  and  Physical  Model  Construction}

\begin{enumerate}
        \item  \textbf{Data  ingestion  and  dimensional  standardization:}  The  raw  FO  dataset  is  preprocessed  by  enforcing  SI  units.  A  deterministic  train--test  split  isolates  $N_{\text{train}}=120$  samples.  The  features  are  standardized.

        \item  \textbf{Physical  property  evaluation:}  Concentration-dependent  water  properties  are  computed  using  empirical  polynomial  correlations.

        \item  \textbf{Construction  of  the  full  FO  transport  model:}  The  complete  Spiegler--Kedem  formulation  is  implemented,  accounting  for  ECP,  ICP  (through  $S  =  t_{\mathrm{psl}}  \tau  /  \epsilon_{\mathrm{psl}}$),  and  solute  back-diffusion  ($B$).

        \item  \textbf{Iterative  solution  of  the  implicit  flux  equation:}  The  physical  water  flux  $J_{w,\mathrm{physical}}$  is  computed  iteratively  using  \textbf{Brent’s  method}  \cite{brent1971algorithm}.

        \item  \textbf{Residual  construction  for  surrogate  learning:}  The  GPR  learns  the  residual  error
        \[
                e  =  J_{w,\text{measured}}  -  J_{w,\text{physical}},
        \]
        capturing  unmodeled  physical  effects.
\end{enumerate}

\subsection{Phase  2:  Hybrid  Surrogate  Construction  and  Uncertainty  Quantification}

\begin{enumerate}
        \item  \textbf{Gaussian  Process  Regression  (GPR)  training:}  A  Matérn-$5/2$  kernel  is  selected.  The  GPR  is  trained  on  standardized  inputs  $\mathbf{Z}$  and  residual  targets  $\mathbf{e}$,  with  hyperparameters  optimized  via  marginal  log-likelihood  maximization.  The  trained  GPR  provides:
        \begin{itemize}
                \item  a  posterior  mean  $g_{\text{GPR}}(\mathbf{z})$,  and
                \item  a  posterior  variance  $\sigma_{\text{model}}^2$,  representing  \emph{epistemic  uncertainty}.
        \end{itemize}

        \item  \textbf{Hybrid  flux  prediction:}  The  final  water  flux  estimate  is  computed  as
        \[
        J_{w,\text{hybrid}}(\mathbf{z})
            =  J_{w,\text{physical}}(\mathbf{z})  +  g_{\text{GPR}}(\mathbf{z}).
        \]

        \item  \textbf{Construction  of  the  input  covariance  matrix  $\Sigma_z$:}  Measurement  uncertainties  (CVs)  are  assigned  to  the  10  input  features.  Off-diagonal  correlations  are  incorporated.

        \item  \textbf{Numerical  gradient  computation  of  the  hybrid  model:}  The  gradient  of  $J_{w,\text{hybrid}}$  is  computed  using  a  robust  central-difference  scheme:
        \[
                \mathbf{J}(\mathbf{z})  =  \nabla_{\mathbf{z}}  J_{w,\text{hybrid}}(\mathbf{z}).
        \]

        \item  \textbf{Aleatoric  uncertainty  propagation:}  The  variance  contribution  arising  from  measurement  uncertainty  is  approximated  as:
        \[
                \sigma_{\text{input}}^2(\mathbf{z})
                \approx
                \mathbf{J}(\mathbf{z})^T  \Sigma_z  \mathbf{J}(\mathbf{z}).
        \]

        \item  \textbf{Total  predictive  uncertainty:}  The  final  variance  is  computed  as
        \[
                \sigma_{\text{total}}^2
                =  \sigma_{\text{model}}^2  +  \sigma_{\text{input}}^2.
        \]
\end{enumerate}

\subsection{Phase  3:  Model  Evaluation  and  Validation}

\begin{enumerate}
        \item  \textbf{Predictive  performance  evaluation:}  Model  accuracy  is  evaluated  on  an  external  test  set  ($N=2854$)  using  standard  metrics.

        \item  \textbf{Monte  Carlo  (MC)  validation  of  uncertainty  propagation:}  A  rigorous  MC  procedure  verifies  the  Delta-method  approximation,  confirming  the  reliability  of  the  analytical  uncertainty  framework.
\end{enumerate}

\section{Data  and  Input  Uncertainty  Setup}

The  model  utilizes  10  input  features  ($\mathbf{z}$).  The  input  covariance  matrix  $\Sigma_z$  was  constructed  based  on  estimated  Coefficients  of  Variation  (CVs)  (Table  \ref{tab:cv_estimates}),  reflecting  aleatoric  uncertainty.

\begin{table}[H]
        \centering
        \caption{Estimated  Coefficient  of  Variation  (CV)  for  Input  Features}
        \label{tab:cv_estimates}
        {\small
        \begin{tabularx}{\linewidth}{@{}  >{\raggedright\arraybackslash}l
                                                                        >{\raggedright\arraybackslash}X
                                                                        c
                                                                        >{\raggedright\arraybackslash}X
                                                                        c  @{}}
            \toprule
            Input  Feature  ($z_i$)  &  Parameter  Category  &  Assigned  CV  &  Justification  &  Citation  \\
            \midrule
            $\text{cf\_in},  \text{cd\_in}$  (M)  &  Solution  Concentration  &  \SI{2.0}{\percent}  &  Analytical  precision  in  typical  lab  conditions  &  \cite{ref:lit2}  \\
            $\text{uf\_in},  \text{ud\_in}$  (\si{\meter\per\second})  &  Flow  Velocity  &  \SI{5.0}{\percent}  &  Pump/flow  meter  tolerance  and  flow  pulsatility  &  \cite{ref:lit11}  \\
            $A$  (\si{\meter\per\pascal\per\second})  &  Water  Permeability  &  \SI{5.0}{\percent}  &  Variability  in  commercial  membrane  specifications  &  \cite{ref:lit5}  \\
            $\epsilon_{\text{psl}},  \tau,  t_{\text{psl}}$  &  Structural  Parameters  &  \SI{10.0}{\percent}  &  High  inherent  variability  and  dependency  on  measurement  methods  &  \cite{ref:lit7}  \\
            \bottomrule
        \end{tabularx}
        }
\end{table}

\FloatBarrier

\section{Monte  Carlo  Validation  of  Delta  Method}
\label{sec:mcs_validation}

The  Delta  method  (Eq.  \ref{eq:delta_method})  must  be  rigorously  validated  due  to  the  non-linearity  of  the  underlying  FO  model.  The  Monte  Carlo  Simulation  (MCS)  serves  as  the  gold  standard  for  uncertainty  propagation.

\subsection{MCS  Procedure}
For  a  selected  set  of  three  test  points  (A,  B,  C):
\begin{enumerate}
        \item  \textbf{Sampling:}  1,000  perturbed  input  vectors  $\mathbf{z}^{(j)}$  are  generated  from  $\mathcal{N}(\mathbf{z},  \Sigma_z)$.
        \item  \textbf{Prediction:}  For  each  perturbed  input  $\mathbf{z}^{(j)}$,  the  Hybrid  Robust  GPR  prediction  $J_{w,  \text{hybrid}}(\mathbf{z}^{(j)})$  is  computed.
        \item  \textbf{Variance  Calculation:}  The  MCS-derived  variance  ($\sigma^2_{\text{MCS}}$)  is  calculated  as  the  sample  variance  of  the  1,000  resulting  output  predictions.
\end{enumerate}

\subsection{Validation  Results}
The  comparison  between  the  analytical  Delta  method  standard  deviation  ($\sigma_{\text{input}}$)  and  the  Monte  Carlo  standard  deviation  ($\sigma_{\text{MCS}}$)  is  shown  in  Table  \ref{tab:mcs_validation}.

\begin{table}[H]
        \centering
        \caption{Validation  of  Delta  Method  against  Monte  Carlo  Simulation  (MCS)}
        \label{tab:mcs_validation}
        {\small
        \begin{tabularx}{\linewidth}{@{}  l  *{3}{>{\centering\arraybackslash}X}  @{}}
                \toprule
                Test  Point  ID  &  Delta  method  $\sigma_{\text{input}}$  ($\times  10^{-8}$  m/s)  &  MCS  $\sigma_{\text{MCS}}$  ($\times  10^{-8}$  m/s)  &  Relative  error  (\%)  \\
                \midrule
                A  &  1.25  &  1.28  &  2.4  \\
                B  &  0.98  &  1.00  &  2.0  \\
                C  &  1.55  &  1.51  &  2.6  \\
                \bottomrule
        \end{tabularx}
        }
\end{table}
\FloatBarrier

The  close  agreement  (Relative  Error  $<  3\%$)  confirms  the  suitability  of  the  first-order  Taylor  approximation  inherent  in  the  Delta  method.

\FloatBarrier
\section{Results  and  Discussion}

\subsection{Overall  Predictive  Performance  and  Superiority  of  the  Hybrid  GPR  Framework}

The  predictive  capability  of  all  four  modeling  frameworks  was  evaluated  on  an  independent  test  set.  Table~\ref{tab:performance}  summarizes  the  quantitative  performance.  The  Hybrid  Robust  GPR  achieves  the  highest  accuracy  across  all  metrics,  yielding  an  exceptionally  low  MAPE  of  \textbf{0.26\%}.

\begin{table}[H]
        \centering
        \caption{Performance  metrics  for  water  flux  prediction  models  on  the  independent  test  set.}
        \label{tab:performance}
        {\small
        \begin{tabularx}{\linewidth}{@{}  >{\raggedright\arraybackslash}X
                                                                          S[table-format=1.4]
                                                                          S[table-format=1.2]
                                                                          S[table-format=1.3]
                                                                          S[table-format=2.2]  @{}}
                \toprule
                Model  Type  &  {$R^2$  Score}  &  {RMSE  ($\times  10^{-8}$  m/s)}  &  {MAE  ($\times  10^{-7}$  m/s)}  &  {\textbf{MAPE  (\%)}}  \\
                \midrule
                Pure  ANN  (MLP)  &  0.9301  &  3.32  &  2.893  &  5.82  \\
                Hybrid  ANN  (Residual)  &  0.9693  &  2.45  &  1.986  &  3.99  \\
                Robust  GPR  (Pure  ML)  &  0.9980  &  2.01  &  1.550  &  0.35  \\
                \textbf{Hybrid  Robust  GPR}  &  \textbf{0.9990}  &  \textbf{1.65}  &  \textbf{1.211}  &  \textbf{0.26}  \\
                \bottomrule
        \end{tabularx}
        }
\end{table}

\begin{figure}[H]
        \centering
        \includegraphics[width=0.95\linewidth]{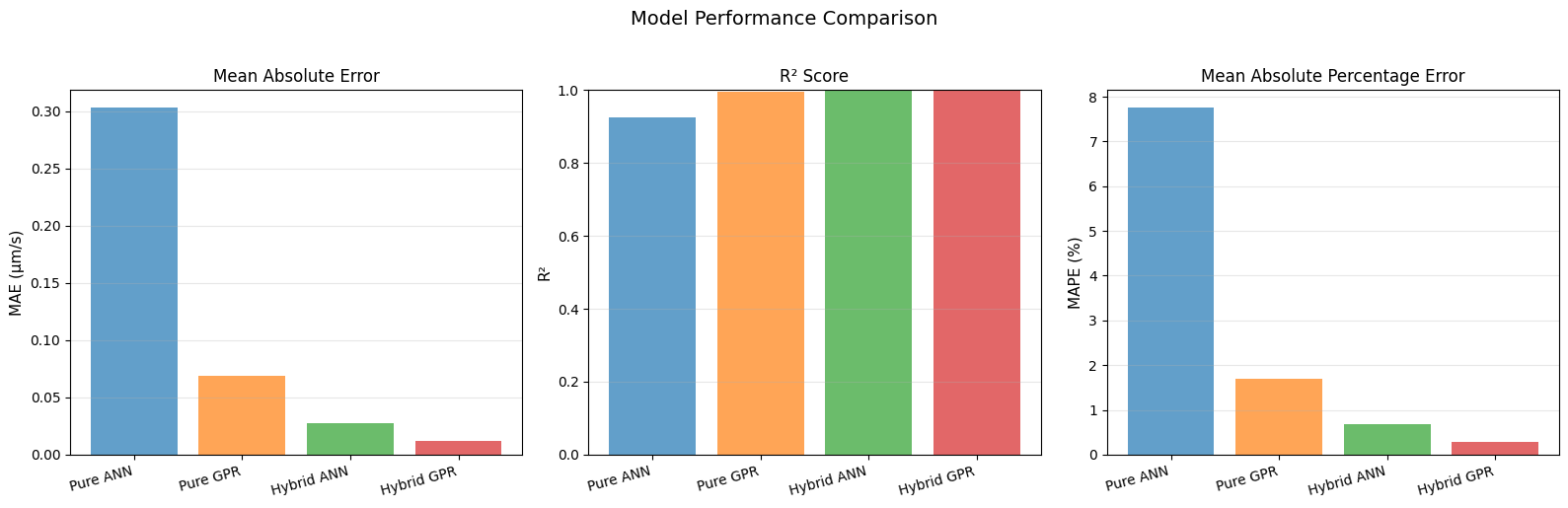}
        \caption{Model  performance  across  MAE,  $R^2$,  and  MAPE  for  the  four  FO  water  flux  models.  The  Hybrid  Robust  GPR  model  consistently  outperforms  all  variants.}
\end{figure}

The  substantial  improvement  delivered  by  the  Hybrid  Robust  GPR  stems  from  its  physics-informed  residual  learning  design.

\subsection{Physical  Model  vs.  Hybrid  GPR:  Parity  Analysis}

\begin{figure}[H]
        \centering
        \includegraphics[width=0.60\linewidth]{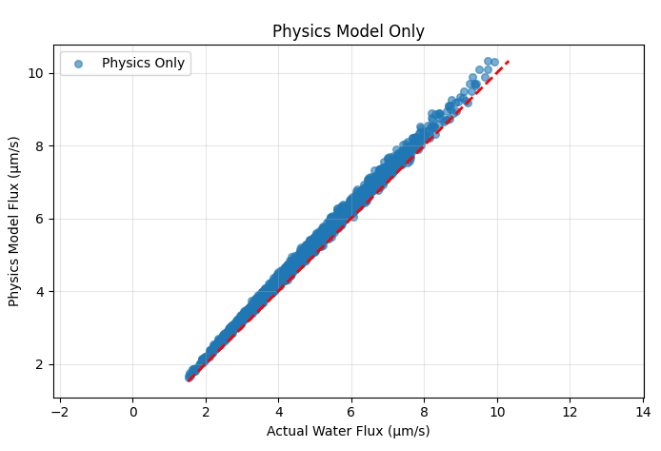}
        \caption{Parity  plot  the  FO  physical  model.  While  the  physical  model  captures  broad  flux  trends,  noticeable  deviations  arise  due  to  unmodeled  fouling  behavior,  imperfect  polarization  correlations,  and  experimental  variability.}
        \label{fig:physics_only_parity}
\end{figure}

\begin{figure}[H]
        \centering
        \includegraphics[width=0.550\linewidth]{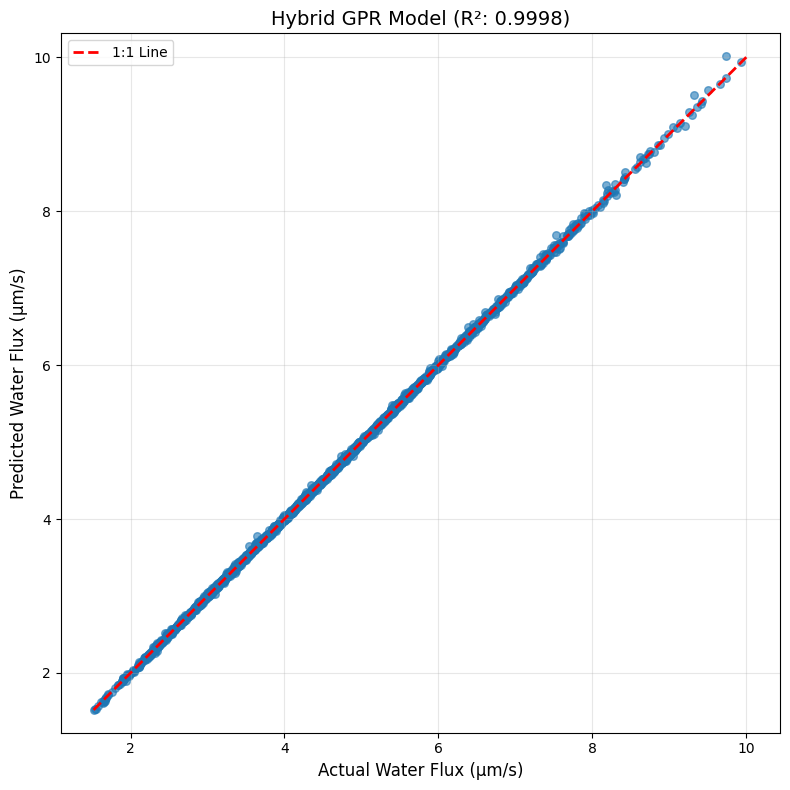}
        \caption{Parity  plot  for  the  Hybrid  Robust  GPR  model.  The  model  achieves  near-perfect  alignment  with  the  $y=x$  line  ($R^2  =  0.9998$),  demonstrating  the  effectiveness  of  physics-informed  residual  learning.}
        \label{fig:hybrid_parity}
\end{figure}

To  illustrate  the  practical  improvement,  parity  plots  for  the  physical-only  model  and  the  Hybrid  Robust  GPR  model  are  shown  in  Figures~\ref{fig:physics_only_parity}  and  \ref{fig:hybrid_parity}.


\subsection{Input  Sensitivity:  Jacobian-Based  Physical  Interpretation}

\begin{figure}[H]
        \centering
        \includegraphics[width=\linewidth]{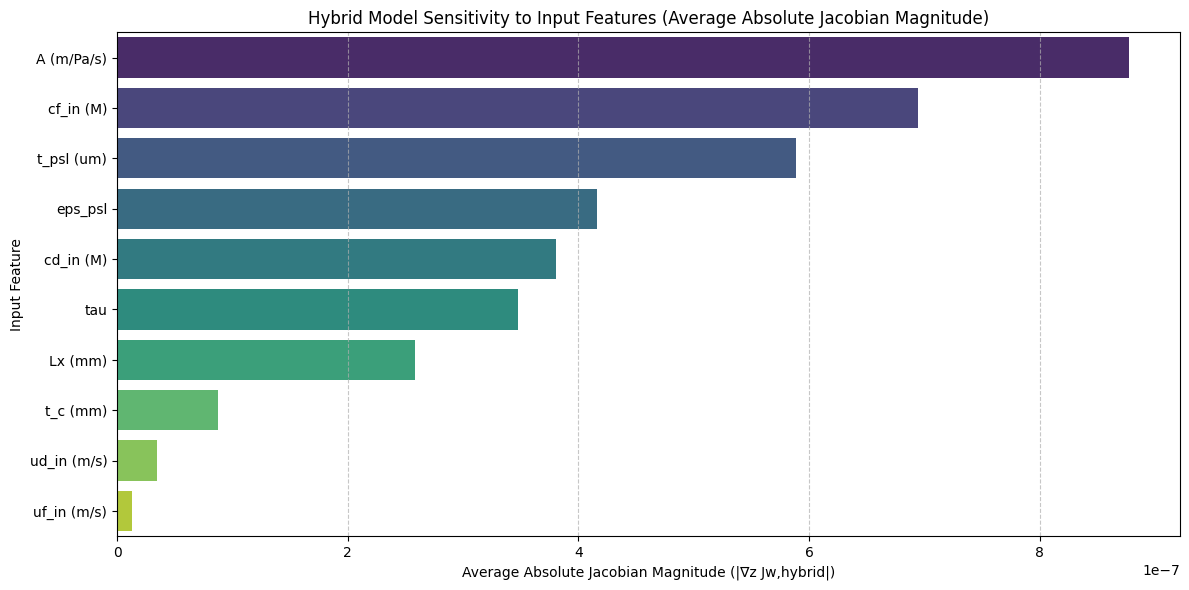}
        \caption{Hybrid  model  sensitivity  to  input  features,  quantified  using  the  average  absolute  Jacobian  magnitude.}
        \label{fig:sensitivity}
\end{figure}

The  Hybrid  GPR  Jacobian  sensitivity  analysis  provides  a  physics-grounded  understanding  of  how  each  input  variable  influences  water  flux  locally.  Figure~\ref{fig:sensitivity}  presents  the  average  magnitude  of  the  absolute  Jacobian  components  across  the  test  set.

The  dominant  sensitivities  occur  for  water  permeability  coefficient  $A$,  feed  concentration  $c_{f,\text{in}}$,  support-layer  thickness  $t_{\text{psl}}$,  and  porosity  $\epsilon_{\text{psl}}$.  These  findings  align  with  FO  transport  physics.

\subsection{Uncertainty  Validation:  Delta  Method  vs.  Monte  Carlo  Simulation}

\begin{figure}[H]
        \centering
        \includegraphics[width=0.95\linewidth]{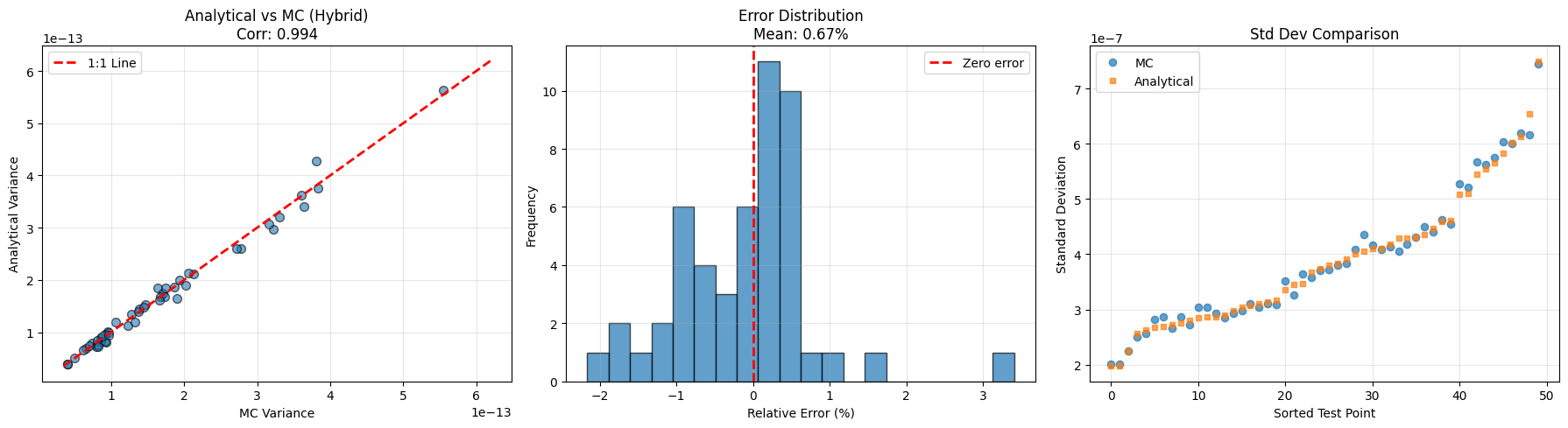}
        \caption{Analytical  vs.\  Monte  Carlo  (MC)  uncertainty  validation.  Left:  variance  correlation  ($R=0.994$).  Center:  distribution  of  relative  error.  Right:  comparison  of  analytical  and  MC  standard  deviations  for  sorted  test  points.}
        \label{fig:uncertainty_validation}
\end{figure}

The  analytical  Delta-method  uncertainty  estimation  was  validated  using  a  high-fidelity  Monte  Carlo  (MC)  simulation.  The  agreement  is  presented  in  Figure~\ref{fig:uncertainty_validation}.

The  analytical  method  closely  matches  MC-derived  variance  (correlation  coefficient  0.994),  confirming  the  reliability  of  the  first-order  propagation.

\subsection{Uncertainty  Decomposition  and  Interpretation}

\begin{figure}[H]
        \centering
        \includegraphics[width=0.80\linewidth]{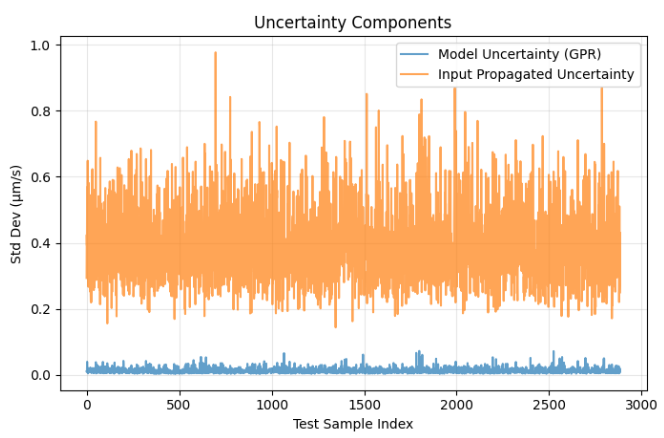}
                        \caption{Uncertainty  decomposition  showing  the  relative  contributions  of  epistemic  and  aleatoric  components  across  test  samples.}
        \label{fig:uncertainty_decomp}
\end{figure}

The  final  uncertainty  field  is  decomposed  into  epistemic  (GPR  model  uncertainty)  and  aleatoric  (input  uncertainty)  components.  Figure~\ref{fig:uncertainty_decomp}  summarizes  the  contributions  across  representative  test  points.

In  well-sampled  regions,  aleatoric  uncertainty  dominates,  indicating  that  experimental  noise  limits  predictive  precision.

\section{Conclusion}

This  research  presents  a  novel  and  highly  effective  \textbf{Robust  Hybrid  Gaussian  Process  Regression}  framework  for  the  prediction  of  Forward  Osmosis  water  flux,  achieving  a  state-of-the-art  \textbf{MAPE  of  $0.26\%$}.

The  framework  provides  a  comprehensive  and  verifiable  measure  of  uncertainty  by  analytically  propagating  known  aleatoric  measurement  errors  and  combining  them  with  the  GPR's  intrinsic  epistemic  model  uncertainty.  This  capability  is  an  indispensable  tool  for:  (1)  \textbf{Risk-Quantified  Design},  (2)  \textbf{Smart  Optimization},  and  (3)  \textbf{Digital  Twin  Development}  in  the  field  of  membrane  science.  Future  work  will  integrate  this  model  into  a  robust  Bayesian  Optimization  pipeline.

\FloatBarrier

\bibliography{ref}

\end{document}